%% file: main.tex
\documentclass{article}

\usepackage[preprint, nonatbib]{neurips_2023}

\usepackage[utf8]{inputenc} %
\usepackage[T1]{fontenc}    %
\usepackage{url}            %
\usepackage{booktabs}       %
\usepackage{amsfonts}       %
\usepackage{nicefrac}       %
\usepackage{microtype}      %
\usepackage{arydshln}
\usepackage[dvipsnames]{xcolor}
\usepackage[pagebackref=true,breaklinks=true,letterpaper=true,colorlinks,bookmarks=false, citecolor=ForestGreen]{hyperref}
\usepackage{times}
\usepackage{epsfig}
\usepackage{graphicx}
\usepackage{amsmath,bm}
\usepackage{amssymb}
\usepackage{algorithmic}
\usepackage[ruled,vlined]{algorithm2e}
\usepackage{multirow}
\usepackage{multicol} 
\usepackage{array}
\usepackage{makecell}
\usepackage{subfigure}
\usepackage{wrapfig}

\usepackage{listings}

\definecolor{codegreen}{rgb}{0,0.6,0}
\definecolor{codegray}{rgb}{0.5,0.5,0.5}
\definecolor{codepurple}{rgb}{0.58,0,0.82}
\definecolor{backcolour}{rgb}{0.95,0.95,0.92}

\lstdefinestyle{mystyle}{
    backgroundcolor=\color{backcolour},   
    commentstyle=\color{codegreen},
    keywordstyle=\color{magenta},
    numberstyle=\tiny\color{codegray},
    stringstyle=\color{codepurple},
    basicstyle=\ttfamily\footnotesize,
    breakatwhitespace=false,         
    breaklines=true,                 
    captionpos=b,                    
    keepspaces=true,                 
    numbers=left,                    
    numbersep=5pt,                  
    showspaces=false,                
    showstringspaces=false,
    showtabs=false,                  
    tabsize=2
}

\usepackage[capitalize]{cleveref}
\crefname{section}{Sec.}{Secs.}
\Crefname{section}{Section}{Sections}
\Crefname{table}{Table}{Tables}
\crefname{table}{Tab.}{Tabs.}

	\newcommand{\qileft}{[\kern-0.15em[}
	\newcommand{\qiLeft}{\left[\kern-0.4em\left[}
	\newcommand{\qiright}{]\kern-0.15em]}
	\newcommand{\qiRight}{\right]\kern-0.4em\right]}

	\renewcommand{\Roman}[1]{\uppercase\expandafter{\romannumeral#1}}

\title{Can GPT-4 Perform Neural Architecture Search?}

\author{
Mingkai Zheng$^{1,3}$ \quad Xiu Su$^{1}$ \quad Shan You$^{2}$ \quad Fei Wang$^{2}$ \\
\textbf{Chen Qian}$^{2}$ \quad \textbf{Chang Xu}$^{1}$ \quad \textbf{Samuel Albanie}$^{3}$  \\
$^1$The University of Sydney ~
$^2$SenseTime Research ~
$^3$CAML Lab, University of Cambridge \\
{\tt\small mingkaizheng@outlook.com},~ {\tt\small xisu5992@uni.sydney.edu.au,} \\ {\tt\small \{youshan,wangfei,qianchen\}@sensetime.com},~ {\tt\small c.xu@sydney.edu.au} \\ {\tt\small samuel.albanie.academic@gmail.com}
 \\ 
}

\begin{document}

\maketitle

\begin{abstract}
  We investigate the potential of GPT-4~\cite{gpt4} to perform Neural Architecture Search (NAS)---the task of designing effective neural architectures. 
Our proposed approach, \textbf{G}PT-4 \textbf{E}nhanced \textbf{N}eural arch\textbf{I}tect\textbf{U}re \textbf{S}earch (GENIUS), leverages the generative capabilities of GPT-4 as a black-box optimiser to quickly navigate the architecture search space, pinpoint promising candidates, and iteratively refine these candidates to improve performance. 
We assess GENIUS across several benchmarks, comparing it with existing state-of-the-art NAS techniques to illustrate its effectiveness.
Rather than targeting state-of-the-art performance, our objective is to highlight GPT-4's potential to assist research on a challenging technical problem through a simple prompting scheme that requires relatively limited domain expertise.\footnote{
Code available at \href{https://github.com/mingkai-zheng/GENIUS}{https://github.com/mingkai-zheng/GENIUS}.}.
More broadly, we believe our preliminary results point to future research that harnesses general purpose language models for diverse optimisation tasks.
We also highlight important limitations to our study, and note implications for AI safety.
\end{abstract}

\input{section/introduction}
\input{section/related_works}

\input{section/method}
\input{section/experiments}

\input{section/imagenet}

\input{section/limitation}

\input{section/safety}

\input{section/conclusion}

\input{section/acknowledgements}

{\small
\bibliographystyle{splncs04}
\bibliography{egbib}
}

\newpage
\appendix
\input{section/appendixv2}

\end{document}

%% file: section/introduction.tex
\section{Introduction}

Recent years have witnessed a string of high-profile scientific breakthroughs by applying deep neural networks to problems spanning domains such as protein folding~\cite{jumper2021highly}, exoplanet detection~\cite{shallue2018identifying} and drug discovery~\cite{stokes2020deep}.
To date, however, successful applications of AI have been marked by the effective use of domain expertise to guide the design of the system, training data and development methodology.

The recent release of GPT-4 represents a milestone in the development of ``general purpose'' systems that exhibit a broad range of capabilities.  
While the full extent of these capabilities remains unknown, preliminary studies and simulated human examinations indicate that the model's knowledge spans many scientific domains~\cite{gpt4,bubeck2023sparks}.
It is therefore of interest to consider the potential for GPT-4 to serve as a general-purpose research tool that substantially reduces the need for domain expertise prevalent in previous breakthroughs.

In this work, we investigate the feasibility of using \mbox{GPT-4} without domain-specific fine-tuning to assist with a research task that has received considerable attention in the machine learning community: deep neural network design.
Deep neural networks have proven effective on a diverse array of language and perception tasks, spanning domains such as question answering~\cite{rajpurkar2016squad}, object recognition \cite{imagenet_cvpr09, cifar} and object detection \cite{pascal-voc-2007, coco}. 
In the quest to improve performance, novel neural architecture designs, exemplified by proposals such as ResNets~\cite{he2016deep} and Transformers~\cite{vaswani2017attention}, have attained substantial gains in performance.
Consequently, there has been significant interest in developing techniques that yield further improvements to neural network architectures.
In particular, \textit{Neural Architecture Search} (\textit{NAS})~\cite{nas1}---which seeks to automate the identification of optimal network configurations for specific tasks---has developed into an active area of research.

The starting point for this work was a pilot exploration that sought to assess whether GPT-4 possessed knowledge relevant for our target task, focusing specifically on the design of Convolution Neural Networks (CNNs).
We began by simply prompting GPT-4\footnote{
\textit{ChatGPT Mar 23 Version.}
}:
\textbf{\textit{``What are the design principles for a high-performance CNN model? Can you provide some examples based on your knowledge?"}}
GPT-4 responded as follows: \textcolor{gray}{\textbf{(1) Depth}: Deeper networks can learn complex features and patterns. Increasing depth by adding more convolutional layers can improve performance while controlling overfitting. \textbf{(2) Width}: Increasing the number of filters in convolutional layers can help capture more diverse and complex features. Striking a balance between width and computational cost is crucial. \textbf{(3) Skip connections}: Skip connections allow gradients to flow through the network more easily, mitigating the vanishing gradient problem and enabling the training of deeper networks. \textbf{(4) Efficient computation}: Techniques like depthwise separable convolutions or group convolutions can reduce the number of parameters and computational costs without sacrificing performance. \textbf{(5) Normalization}: Normalization methods like batch normalization, layer normalization, or instance normalization can improve convergence and stability. }

While the  response to our rudimentary prompt lacks fine-grained specificity, it nevertheless suggests that GPT-4 possesses knowledge relevant to the target task.
A natural next question then arises: does GPT-4 simply have ``surface level'' awareness of information related to this task, or does it have the depth of knowledge required to play a substantially greater role in the design and engineering of neural network architectures?

Our investigation seeks to address this question by examining the degree to which GPT-4 can propose effective neural network architecture designs. %
We view our work as a tentative exploration of the potential of GPT-4 to assist with scientific discovery, providing suggestions that enable rapid research prototyping on a challenging optimisation task.
Further, we suggest that evidence of GPT-4's ability to search neural network architecture design spaces with limited input from human domain experts has implications for AI safety.
However, we also emphasise the preliminary nature of our study and highlight some limitations (\cref{sec:limitations}) to our methodology.

%% file: section/related_works.tex
\section{Related Work}

\subsection{Neural Architecture Design and Search}

Neural architecture design plays a prominent role in deep learning research, with numerous studies focusing on developing architectural enhancements. 
Seminal works such as LeNet-5~\cite{lenet}, AlexNet~\cite{alexnet},~VGGNet~\cite{vgg}, GoogleNet~\cite{googlenet}, ResNet~\cite{resnet}, DenseNet~\cite{densenet}, SENet~\cite{Hu2017SqueezeandExcitationN} and Transformers~\cite{vaswani2017attention} contributed design insights to improve performance.
Numerous subsequent studies \cite{mbv1, mbv2, mbv3, shufflenetv1, shufflenetv2, resnext, densenet, resnest} have further leveraged hand-crafted designs to explore the space of efficient, more capable architectures.

Neural Architecture Search (NAS) builds on many of these ideas but seeks a greater level of automation in the design process.
Early efforts~\cite{nas1, nas2} employed reinforcement learning to explore the search space of potential architectures, with later approaches leveraging evolutionary strategies~\cite{real2017large} and Bayesian optimisation~\cite{kandasamy2018neural}. 
There has been considerable focus on reducing the computational burden associated with the search, with proposals such as DARTS~\cite{darts} leveraging gradient-based search and EfficientNAS~\cite{enas} employing sub-network sampling to increase efficiency.
A rich body of work has further explored this direction~\cite{mcts, fairnas, bignas, greedynas, greedynasv2, kshotnas, proxylessnas, mnasnet, singlepath, efficientnet, vitas, bignas}.
More recent work has employed evolutionary prompt engineering with soft prompt tuning to use language models for evolutionary NAS~\cite{chen2023evoprompting}.
In contrast to conventional search strategies, we employ a process that simply prompts GPT-4 to propose designs from a given search space with a handful of examples.

\subsection{Exploring GPT-4's research capabilities}

Early studies in the technical report accompanying the release of GPT-4~\cite{gpt4} demonstrated that the model can achieve strong results across a broad suite of examinations designed to test human knowledge in widely-studied scientific disciplines such as biology, chemistry, physics, and computer science \cite{bordt2023chatgpt} etc.\footnote{We note that these results should be interpreted cautiously since the tests were designed for humans rather than language models.
Nevertheless, they indicate some degree of familiarity with concepts that form prerequisites for various domains of scientific research.}
A complementary set of preliminary qualitative studies conducted on an early variant of GPT-4 further highlight its ability to perform sophisticated reasoning across many topic areas~\cite{bubeck2023sparks}, a further key building block for research applications.
These studies also note important limitations in the model of relevance for research tasks - these include longstanding problems with ``hallucinations''~\cite{maynez2020faithfulness} and bias~\cite{hovy2016social,gpt3}, as well as an inability to construct appropriate plans in arithmetic and reasoning problems.
Exploring applications in the chemistry domain, concurrent work explores how GPT-4 can be coupled to existing tools like web search and code execution to advance chemical research~\cite{bran2023chemcrow,boiko2023emergent}.
Focusing on a different domain, we view our work as complementary to these explorations.

%% file: section/method.tex
\begin{figure*}
    \centering
    \includegraphics[width=0.8\linewidth]{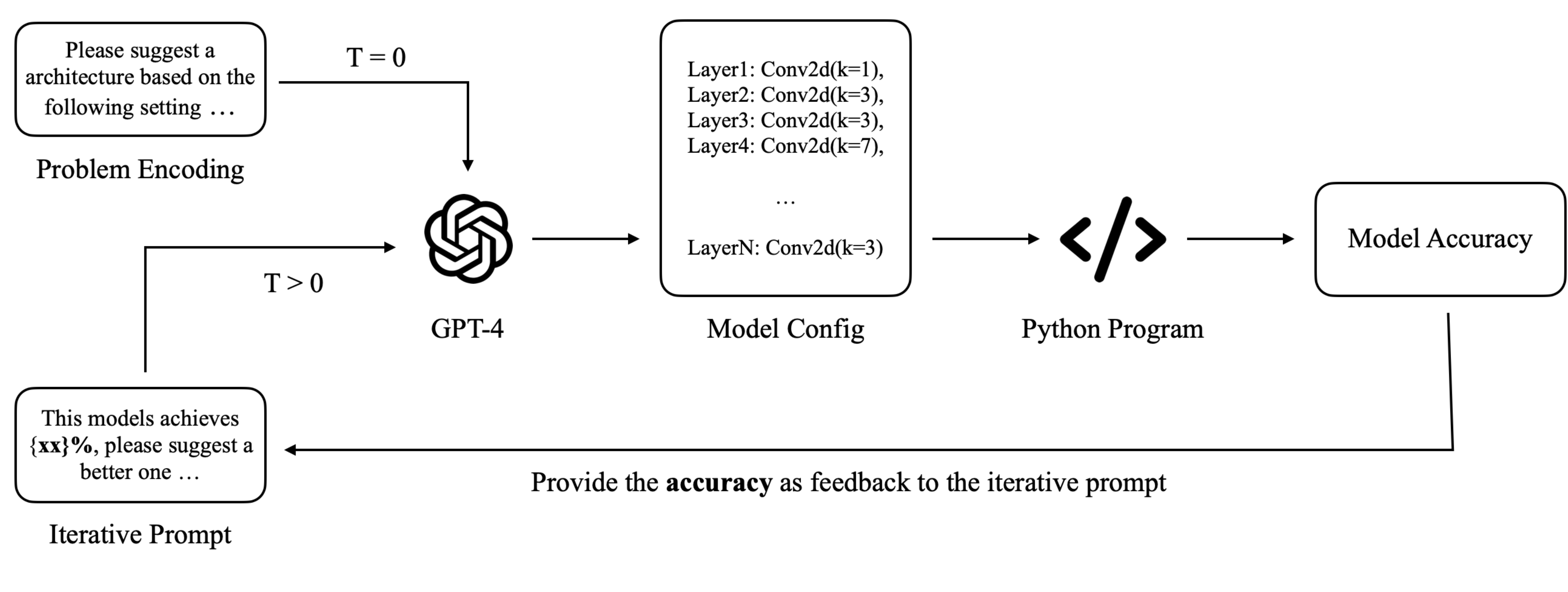}
    \vspace{-10pt}
    \caption{\textbf{An overview of the GENIUS framework}.
    After an initial problem encoding (corresponding to iteration $T = 0$), GPT-4 proposes a model configuration.
    A Python program is then executed to evaluate the quality of the configuration (assessed through its accuracy), and the results are passed back to GPT-4 via a natural language prompt for further iterations.}
    \label{fig:GENIUS}
    \vspace{-10pt}
\end{figure*}

\section{Approach}

Our proposed method, \textbf{G}PT-4 \textbf{E}nhanced \textbf{N}eural arch\textbf{I}tect\textbf{U}re \textbf{S}earch (GENIUS), aims to tackle the challenging neural architecture search (NAS) problem by using GPT-4 as a ``black box'' optimiser. This entails first simply encoding the NAS problem statement into a human-readable text format that GPT-4 can parse. The model then responds with a model configuration proposal that aims to maximise a given performance objective (e.g., accuracy on a particular benchmark). GENIUS operates through an iterative refinement process.
In the first iteration, we provide the problem encoding to the GPT-4 model which responds with an initial model configuration.
Subsequently, we employ training and evaluation code to execute the model and obtain its empirical accuracy.
This performance metric is then passed back to the GPT-4 model, prompting it to generate an improved model based on the insights gained from previous experiments.
The algorithm is depicted in Algorithm~\ref{alg:GENIUS}.

\begin{algorithm}
\SetAlgoLined
\SetKwInOut{Input}{Input}
\Input{\textbf{GPT-4}: The GPT-4 API. \\ \textbf{Problem\_Encoding}: The human-readable text that encodes the NAS problem.\\  \textbf{Run}: The training and testing codes for executing and obtaining the ground truth results.}
\For{T=0 to iteration}{
    \eIf{T == 0}{
        model = GPT-4(Problem\_Encoding)
    } {
        prompt = ``\texttt{By using this model, we achieved an accuracy of \textbf{\{Accuracy\}}\%. Please recommend a new model that outperforms prior architectures based on the abovementioned experiments. Also, Please provide a rationale explaining why the suggested model surpasses all previous architectures.}''
        
        model = GPT-4(prompt)
    }

    \textbf{Accuracy} = Run(model)
}
\SetKwInOut{Output}{Output}
\Output{The Best Model Configuration}
\caption{GPT-4 Enhanced Neural Architecture Search (GENIUS)}
\label{alg:GENIUS}
\end{algorithm}

%% file: section/experiments.tex
\section{Proof of Concept}
In this section, we first apply our GENIUS to two benchmark datasets to validate its effectiveness and empirically investigate its behavior. 
Following this, we assess the performance of the optimal architecture identified by GENIUS on a widely-used benchmark in the NAS domain where we compare to the existing state-of-the-art.

\subsection{Dataset and Benchmark}
1. \textbf{NAS-Bench-Macro}\footnote{\url{https://github.com/xiusu/NAS-Bench-Macro}}  - This benchmark was first proposed in MCT-NAS \cite{mcts} for single-path one-shot NAS methods. It consists of 6561 architectures and their isolated evaluation results on the CIFAR-10 dataset \cite{cifar}. The search space of NAS-Bench-Macro is conducted with 8 searching layers, where each layer contains 3 candidate blocks. These blocks are marked as Identity, InvertedResidual Block with kernel size = 3 and expansion ratio = 3, and InvertedResidual Block with kernel size = 5 and expansion ratio = 6. Thus, the total size of the search space is $3^8 = 6561$.

2. \textbf{Channel-Bench-Macro}\footnote{\url{https://github.com/xiusu/Channel-Bench-Macro}} - This benchmark was first proposed in BCNet \cite{bcnetv2} for channel number search. The search space of this benchmark is conducted with 7 searching layers, where each layer contains 4 uniformly distributed candidate widths. Thus, the overall search space is $4^7 = 16384$. It also provides the test results for all the 16384 architectures on CIFAR10 \cite{cifar}. Additionally, this benchmark includes two base models, MobileNet \cite{mbv2} and ResNet \cite{resnet}.

\subsection{Empirical Study}
\textbf{Random Sampling Baseline.}  In the realm of NAS, randomly sampled architectures are typically employed as a baseline. In the context of this study, we will utilize a stochastic function to uniformly sample from the available operations and channel numbers associated with each layer. Concretely, we will perform 10 sampling iterations and subsequently identify the most optimal architectures to serve as our baselines. Nevertheless, we observed considerable variance across individual trials resulting from this sampling approach. To address this, we repeated the 10-iteration process 10,000 times and calculated the average of the best outcomes.

\begin{figure*}[h]
    \centering
    \subfigure[\footnotesize Accuracy (temp=1) ]{\label{fig:nas_bench_acc_t1}\includegraphics[width=0.24\linewidth]{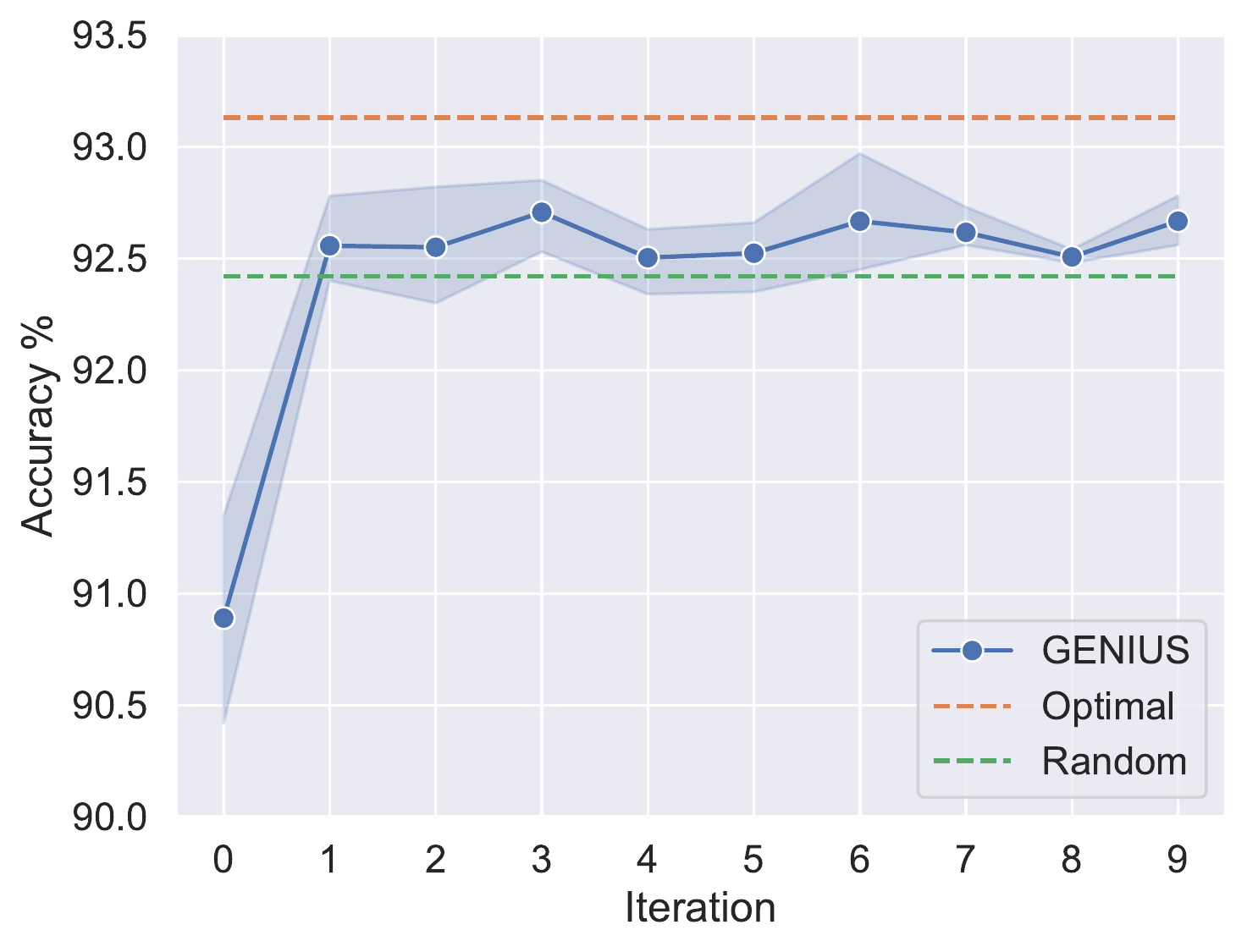}}
    \subfigure[\footnotesize Accuracy (temp=0)]{\label{fig:nas_bench_acc_t0}\includegraphics[width=0.24\linewidth]{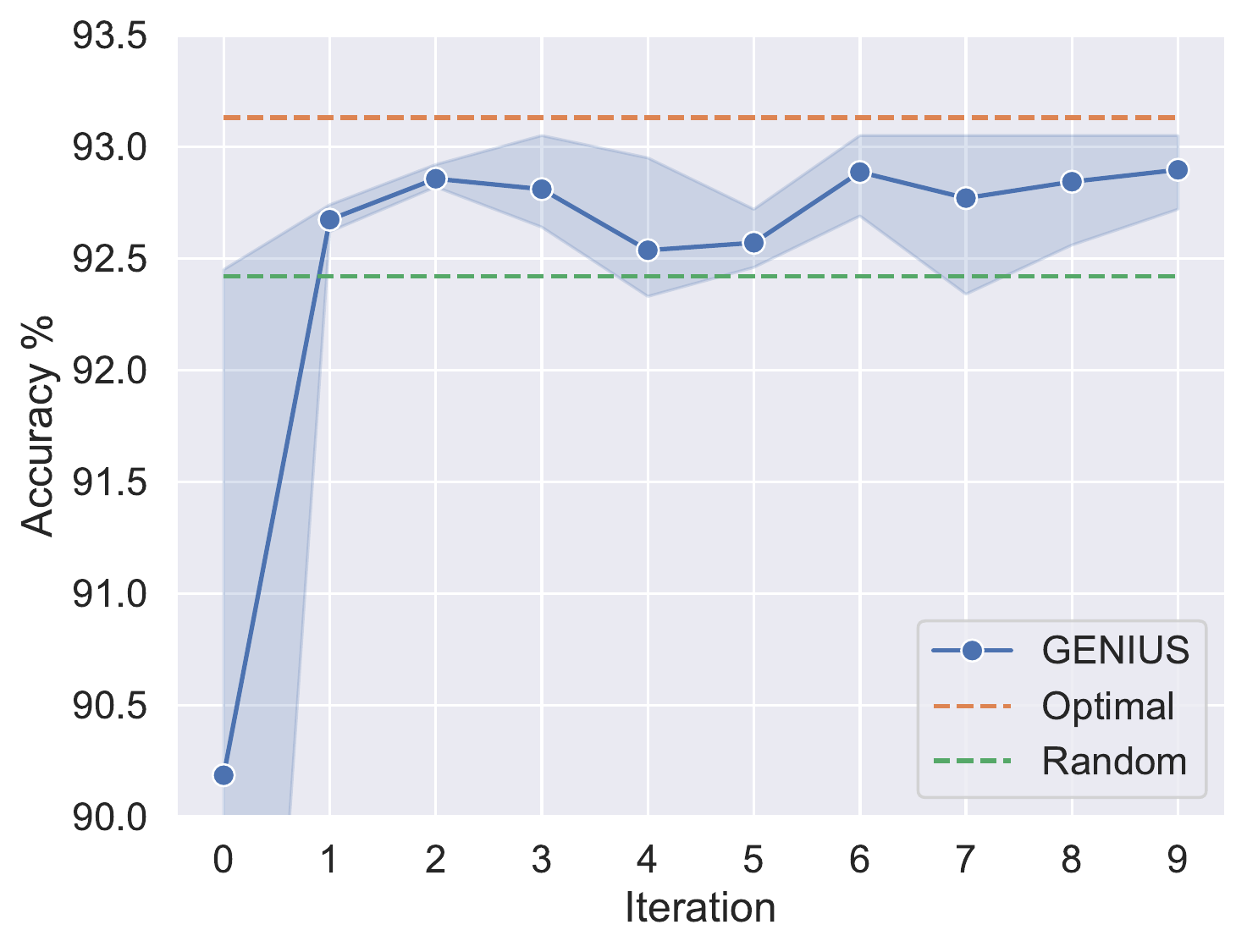}}
    \subfigure[\footnotesize Ranking (temp=1)]{\label{fig:nas_bench_rank_t1}\includegraphics[width=0.24\linewidth]{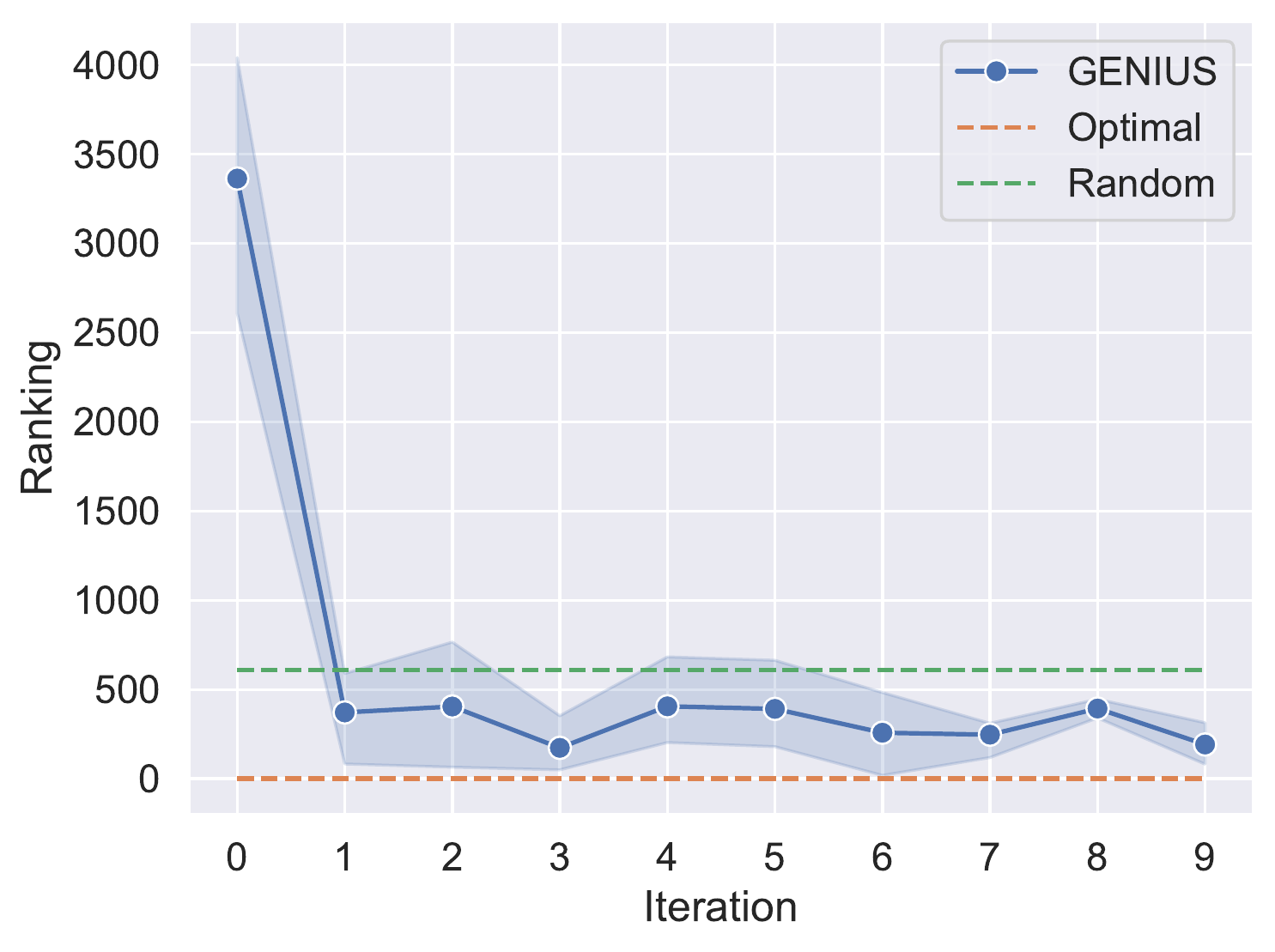}}
    \subfigure[\footnotesize Ranking (temp=0)]{\label{fig:nas_bench_rank_t0}\includegraphics[width=0.24\linewidth]{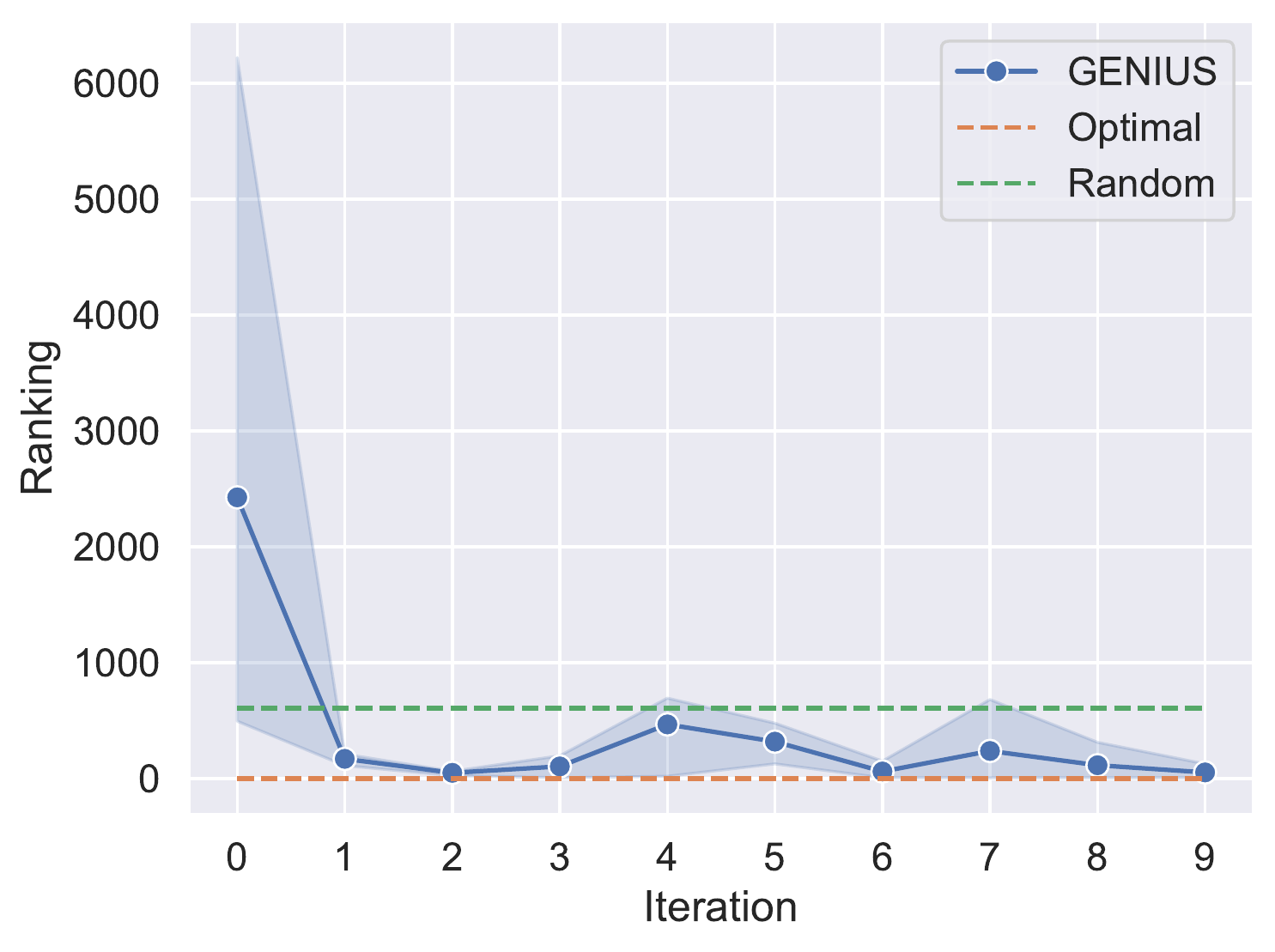}}
    \vspace{-10pt}
    \caption{\footnotesize We conducted experiments on 
    \textbf{NAS-Bench-Macro} and tested the results at two different temperatures: 0 and 1. Each experiment was repeated 3 times with 10 iterations per experiment. We show both the accuracy and ranking for each iteration. It's important to note that higher accuracy and lower ranking numbers indicate better architecture.}
    \label{fig:nas_bench_macro}
\end{figure*}

\textbf{NAS-Bench-Macro}. To assess the effectiveness of GENIUS, we conduct an experiment using the NAS-Bench-Macro. For this experiment, we set the maximum number of iterations to 10. Since the benchmark provides ground truth accuracy values for each model configuration as a lookup table, we use these to retrieve the relevant accuracy score at each step.
The GPT-4 API includes a \textit{temperature} hyperparameter that controls the randomness of the model's output, with higher values leading to greater randomness in the output. 
We conducted experiments with both temperature=0 and temperature=1 to assess the effectiveness of GENIUS under different levels of randomness.

The experimental results are presented in Figure \ref{fig:nas_bench_macro}. We show both the accuracy and the model's ranking for each iteration. The best model obtained is ranked 8/6561 (Top 0.12\%), while the worst model is ranked 61/6561 (Top 0.93\%), remaining reasonable.
(\emph{We provide detailed numerical results for this experiment in Appendix A.1})
We observe that GENIUS exhibits some randomness in its responses, even when the temperature is set to 0. 
Nonetheless, despite this randomness, satisfactory results are achieved in the majority of cases.

\begin{figure*}[h]
    \centering
    \subfigure[\footnotesize ResNet Accuracy ]{\label{fig:res_acc}\includegraphics[width=0.24\linewidth]{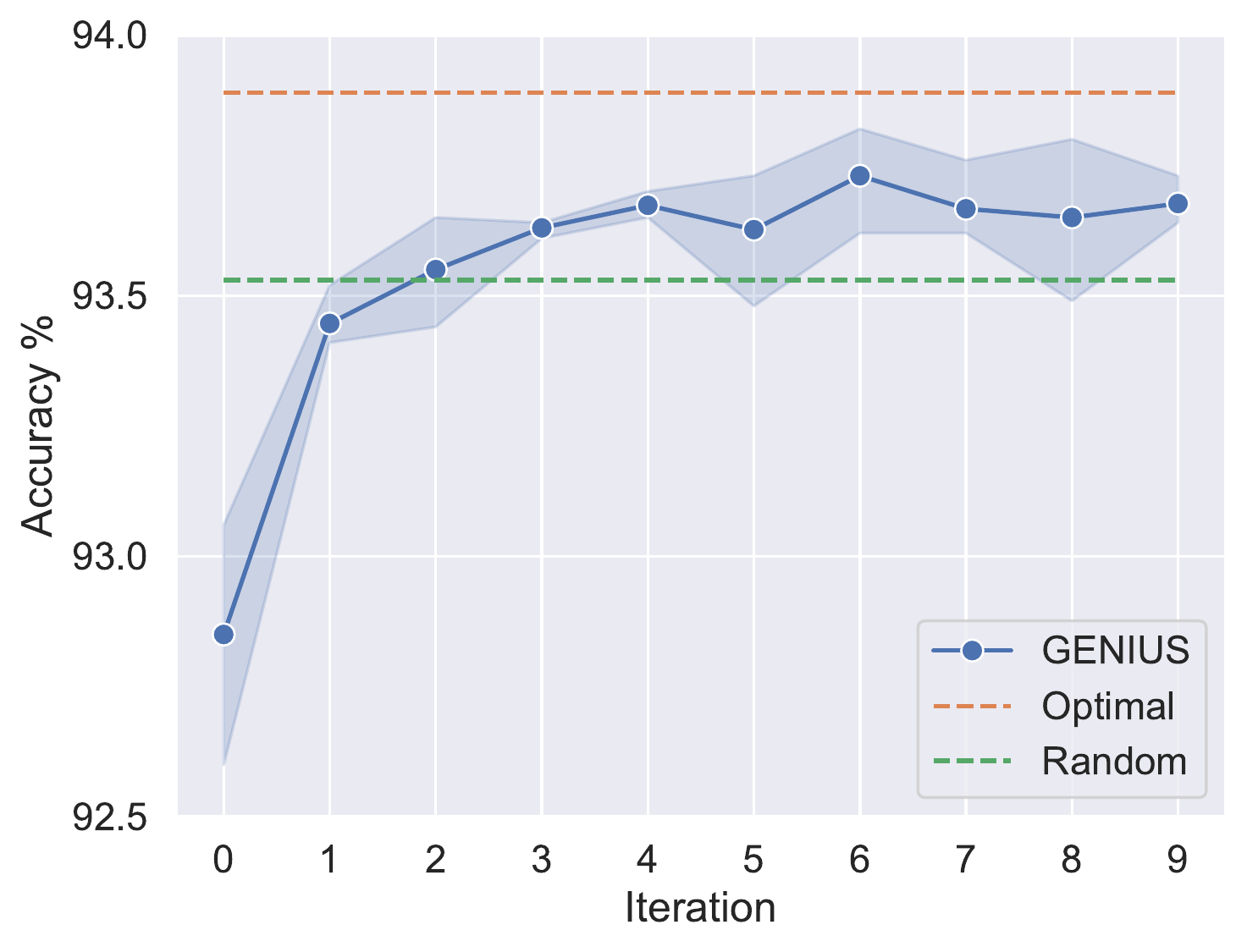}}
    \subfigure[\footnotesize MobileNet Accuracy]{\label{fig:mob_acc}\includegraphics[width=0.24\linewidth]{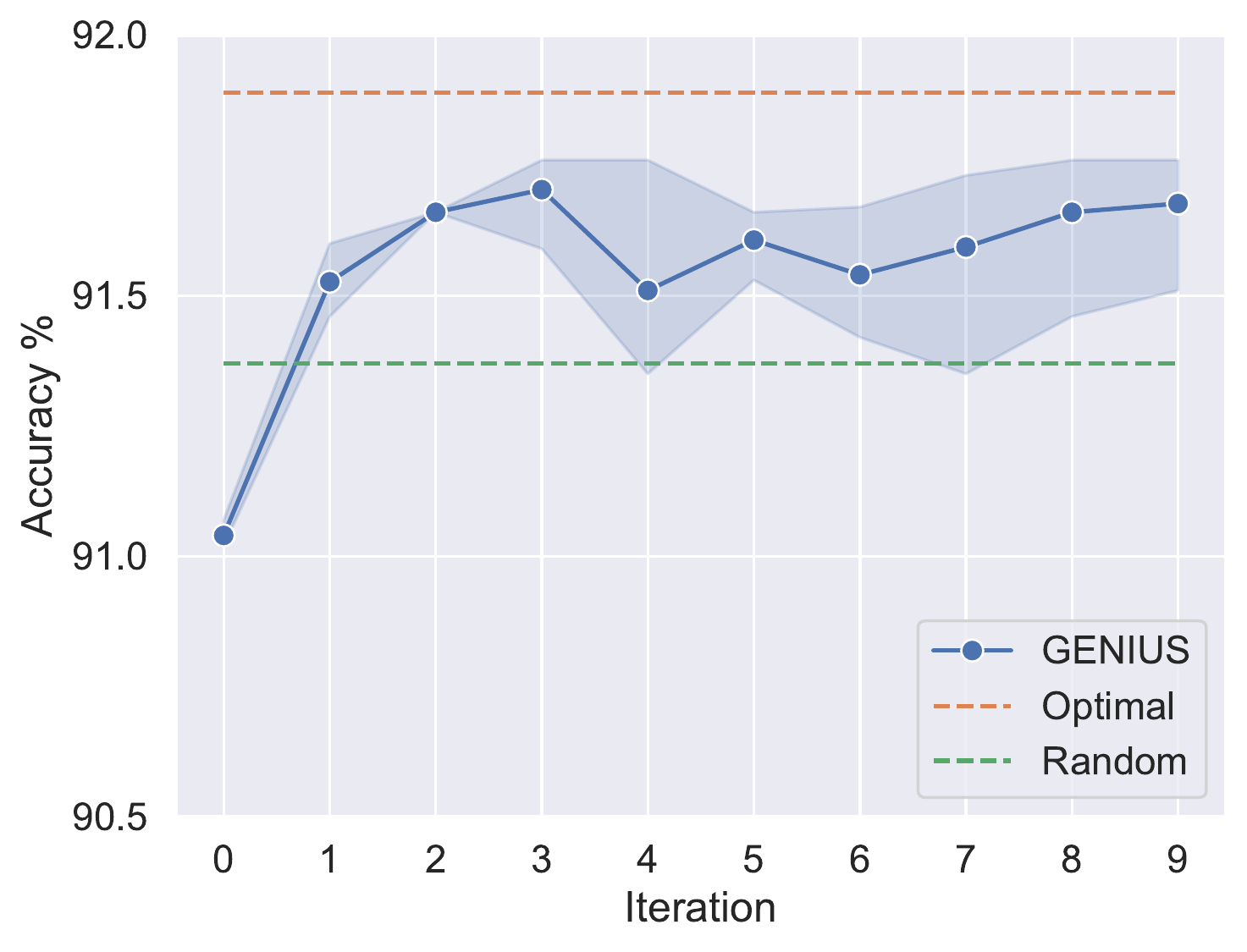}}
    \subfigure[\footnotesize ResNet Ranking]{\label{fig:res_rank}\includegraphics[width=0.24\linewidth]{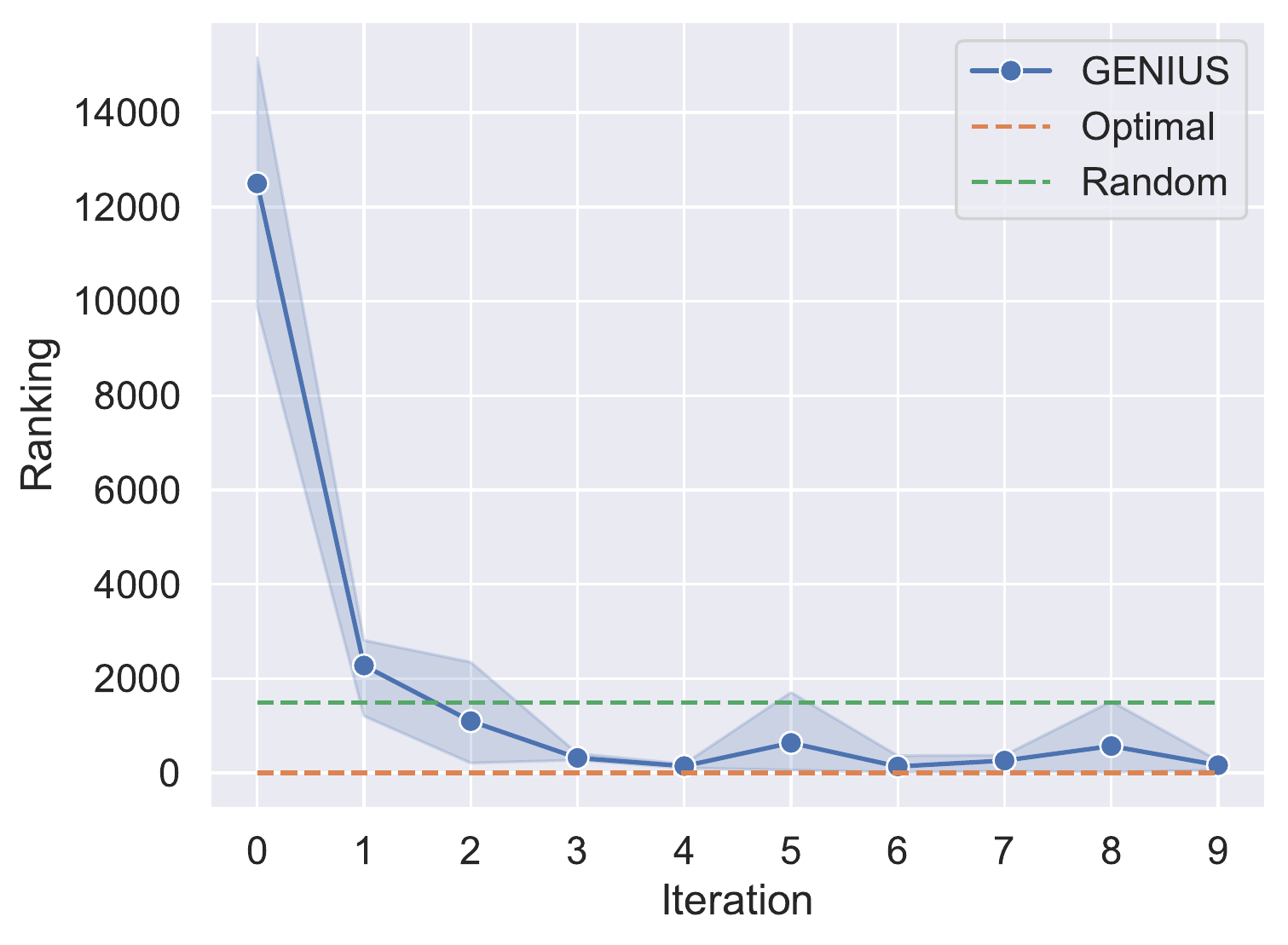}}
    \subfigure[\footnotesize MobileNet Ranking]{\label{fig:mob_rank}\includegraphics[width=0.24\linewidth]{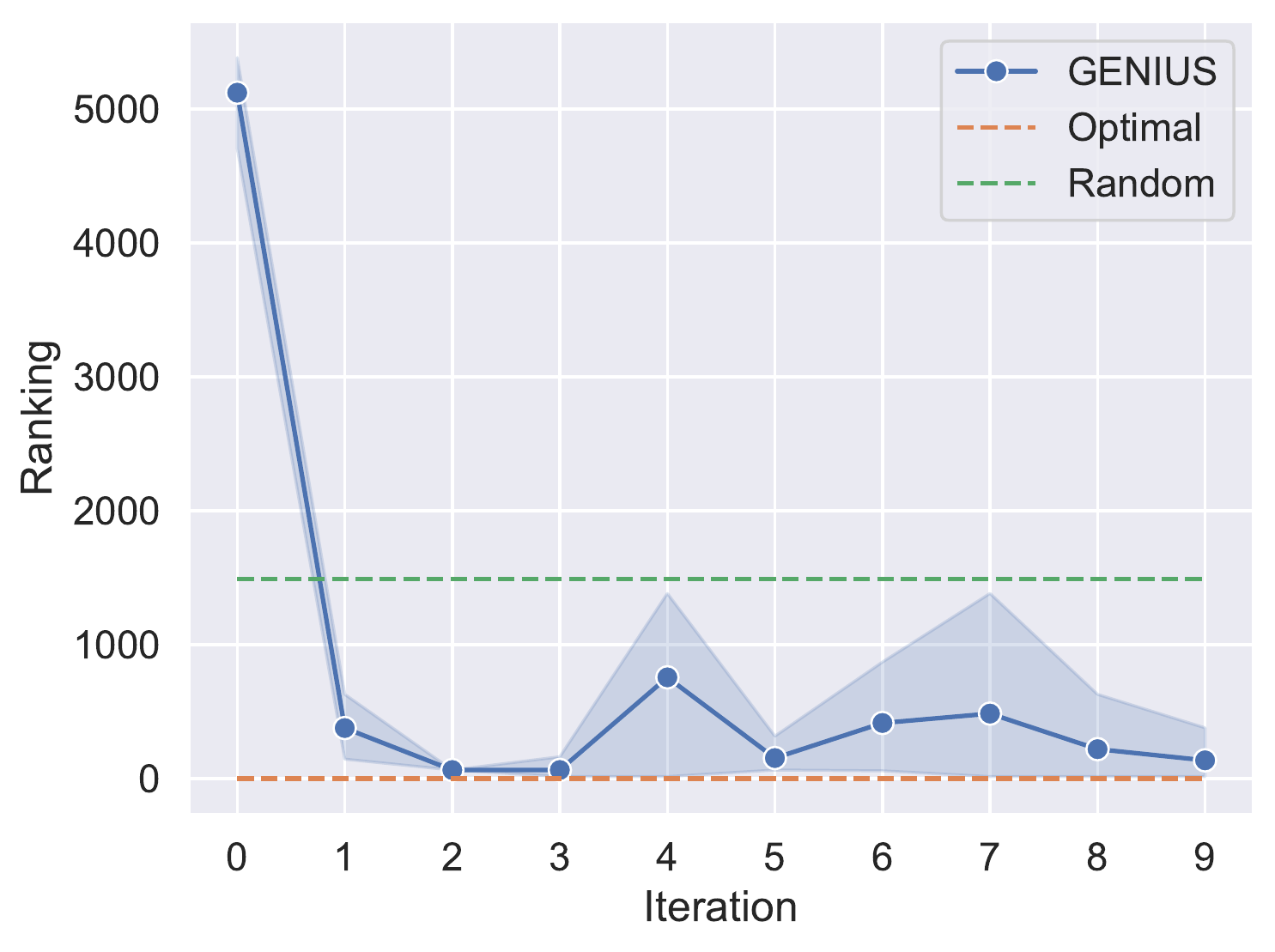}}
    \vspace{-10pt}
    \caption{\footnotesize We performed experiments on the \textbf{Channel-Bench-Macro} benchmark, testing the results on both ResNet and MobileNet base models with a fixed Temperature of 0. Each experiment was conducted 3 times with 10 iterations. We show both accuracy and ranking for each iteration. Similar to previous experiments, higher accuracy and lower ranking numbers indicate better architecture. }
    \label{fig:channel_bench_macro}
\end{figure*}

\textbf{Channel-Bench-Macro.} We further evaluate the effectiveness of GENIUS on Channel-Bench-Macro. In this experiment, we fix the temperature to 0 and perform only one trial on both ResNet and MobileNet settings. The experimental results are presented in Figure \ref{fig:channel_bench_macro}. Similar to the previous experiment, we show the accuracy and rank for 10 iterations. Specifically, GENIUS achieves Rank 33 / 16384 (Top 0.2\%) for the ResNet-based model and Rank 16 / 16384 (Top 0.1\%) for the MobileNet-based model, further demonstrating its effectiveness. (\emph{We provide detailed numerical results for this experiment in Appendix A.2})

\subsection{NAS-Bench-201}

Next, we extend our application of GENIUS to the well-known NAS-Bench-201 \cite{nasbench201} benchmark\footnote{\url{https://github.com/D-X-Y/NAS-Bench-201}}. 
This benchmark focuses on designing a cell block for neural architectures. 
The cell in the search space is represented as a densely connected directed acyclic graph (DAG) consisting of four nodes and six edges, where nodes represent feature maps, and edges correspond to operations. 
With five available operations, the total number of possible search spaces amounts to $5^6 = 15625$. 
We conduct evaluations on CIFAR10, CIFAR100, and ImageNet16-120.

\renewcommand\arraystretch{0.8}
\begin{table}[h]
\centering
\caption{\footnotesize Experimental Results on Nas-Bench-201. We set \textbf{Temperature = 0} for GPT-4 in this experiment. We report the experimental results based on 5 trials for GENIUS. The performances of DRNAS \cite{drnas}, $\beta$-DARTS \cite{betadars}, and $\Lambda$-DARTS \cite{lambdadarts} are identical, potentially attributable to their near-optimal performance on NAS-Bench-201.}
\label{table:channel_bench_201}
\setlength\tabcolsep{3pt}
\footnotesize
\begin{tabular}{l c c c c c c}
        \toprule
        \multirow{2}{*}{Method}  & \multicolumn{2}{c}{CIFAR10} & \multicolumn{2}{c}{CIFAR100} & \multicolumn{2}{c}{ImageNet16-120} \\ 
        & Validation & Test & Validation & Test & Validation & Test  \\ \midrule 
        DARTS \cite{darts}  & 39.77±0.00 &  54.30±0.00 &  38.57±0.00 & 15.61±0.00 &  18.87±0.00 &  16.32±0.00 \\
        DSNAS \cite{dsnas}   & 89.66±0.29 & 93.08±0.13 &  30.87±16.40 & 31.01±16.38 & 40.61±0.09 & 41.07±0.09 \\
        PC-DARTS \cite{pc_darts}  & 89.96±0.15 & 93.41±0.30 & 67.12±0.39 & 67.48±0.89 & 40.83±0.08 & 41.31±0.22 \\
        SNAS \cite{SNAS} & 90.10±1.04 & 92.77±0.83 & 69.69±2.39 & 69.34±1.98 & 42.84±1.79 & 43.16±2.64 \\
        iDARTS \cite{idarts} & 89.86±0.60 & 93.58±0.32 &  70.57±0.24 & 70.83±0.48 &  40.38±0.59 & 40.89±0.68 \\ 
        GDAS \cite{GDAS} & 89.89±0.08 & 93.61±0.09 & 71.34±0.04 & 70.70±0.30 & 41.59±1.33 &  41.71±0.98  \\
        DRNAS \cite{drnas} & 91.55±0.00 & 94.36±0.00  &  73.49±0.00 & 73.51±0.00 & 46.37±0.00 & 46.34±0.00  \\
        $\beta$-DARTS \cite{betadars}  & 91.55±0.00 & 94.36±0.00  & 73.49±0.00 & 73.51±0.00 & 46.37±0.00 & 46.34±0.00  \\
        $\Lambda$-DARTS \cite{lambdadarts} & 91.55±0.00 & 94.36±0.00  &  73.49±0.00 & 73.51±0.00 & 46.37±0.00 & 46.34±0.00  \\ \midrule
        \textbf{GENIUS (Ours)} & 91.07±0.20 & 93.79±0.09 & 70.96±0.33 & 70.91±0.72 & 45.29±0.81 & 44.96±1.02 \\
        \bottomrule
\end{tabular}
\vspace{-3mm}
\end{table}

Consistent with prior experiments, we set the temperature to 0 to minimize randomness and utilize 10 iterations for GENIUS. As this benchmark offers both validation and test accuracy, we employ validation accuracy in the prompt and report the test accuracy corresponding to the highest validation accuracy. 
The results are displayed in Table \ref{table:channel_bench_201}, where we observe that GENIUS achieves competitive performance, though slightly behind leading methods (~\cite{drnas, betadars, lambdadarts}) attaining accuracies of 93.79±0.09, 70.91±0.72, and 44.96±1.02 on CIFAR10, CIFAR100, and ImageNet16-120, respectively. 
(\emph{We also provide additional numerical results for each iteration in Appendix A.3})

%% file: section/imagenet.tex
\section{Large-Scale Experiments}

Our previous experiments demonstrated the potential of using GPT-4 to address the NAS problem. 
However, the three benchmarks \cite{nasbench201, mcts, bcnetv2} considered above have two shortcomings.
First, the search space is limited, ranging from 6561 to 16384 architecture candidates in total.
Second, there is no imposition of a FLOPs constraint, which offers a useful proxy for the computational constraints facing architecture designers who target real-world deployment.
To address these shortcomings, we next evaluate the performance of GENIUS on the ImageNet dataset \cite{imagenet_cvpr09} with the MobileNet V2 search space \cite{mbv2, mbv3}. This setting is widely used for evaluating the performance of NAS algorithms and provides a more realistic and challenging search scenario. 

\textbf{Search Strategy}. 
The ImageNet dataset consists of 1.28 million training images and 50K validation images from 1,000 classes.
To avoid overfitting, we split the training data into 99\% for training and 1\% for validation, retaining the official 50K validation images for testing. 
Following the same protocol as in previous experiments, we run our GENIUS method for 10 iterations and validate its performance on the 1\% validation set. 
Finally, we evaluate the performance of the best architecture selected from the validation set on the 50K testing images.
For each architecture, we train the model for 20 epochs using a standard SGD optimizer with a learning rate of 0.5, weight decay of 1e-4, batch size of 1024, and momentum of 0.9. We use only basic data augmentations such as RandomResizedCrop and RandomFlip during the search stage. We also use 196x196 as the input size to save the search cost.

\textbf{Search Space}. To ensure a fair comparison with recent works \cite{mcts, kshotnas, greedynas}, we conduct architecture search over a search space that includes mobile inverted bottleneck MBConv \cite{mbv2} and squeeze-excitation modules \cite{senet}. The search space comprises seven basic operators, such as MBConv with kernel sizes of {3, 5, 7} and expansion ratios of {4, 6}, as well as a skip connection to enable different depths of architectures. We divide the search space into 5 stages, each containing a maximum of 6 layers. In total, the search space contains approximately $7^{30}$ possible architecture candidates.

\renewcommand\arraystretch{0.8}
\begin{table*}[h]
    \centering
    \setlength\tabcolsep{10pt}
    \caption{\footnotesize Comparison of searched architectures for different NAS methods on ImageNet. Our search cost is measured on V100 GPUs to fair compare with the previous method. $\dagger$ denotes searched on TPUs.}
    \label{exp:imagenet}
    \footnotesize
    \begin{tabular}{l c c c c c c c}
        \toprule
         Method & \thead{FLOPs \\ (M)}  & \thead{Params \\ (M)} & \thead{Top-1 \\ (\%)}  & \thead{Top-5 \\ (\%)} & \thead{Search Cost \\ (GPU Days)} \\ \midrule
         MobileNetV2 \cite{mbv2} & 300 & 3.4 & 72.0 & 91.0 & Human Designed \\
         AngleNet \cite{anglenet} & 325 & - & 74.2 & - & Unkown  \\
         Proxyless-R \cite{proxylessnas} & 320 & 4.0 & 74.6 & 92.2 & 15 \\ 
         MnasNet-A2  \cite{mnasnet} & 340 & 4.8 & 75.6 & 92.7 & 288$\dagger$ \\
         BetaNet-A \cite{betanas} & 333 & 4.1 & 75.9 & 92.8 & 7 \\
         SPOS \cite{singlepath} & 328 & - & 76.2 & - & 12 \\
         SCARLET-B  \cite{scarletnas} & 329 & 6.5 & 76.3 & 93.0 & 22 \\
         ST-NAS-A \cite{st-nas} & 326 & 5.2 & 76.4 & 93.1 & Unkown \\
         GreedyNAS-B \cite{greedynas} & 324 & 5.2 & 76.8 & 93.0 & 7 \\
         MCT-NAS-B \cite{mcts} & 327 & 6.3 & 76.9 & 93.4 & 12  \\
         FairNAS-C \cite{fairnas} & 325 & 5.6 & 76.7 & 93.3 & Unkown \\
         K-shot-NAS-B \cite{kshotnas} & 332 & 6.2 & 77.2 & 93.3 & 12  \\
         FBNetV2-L1 \cite{fbnetv2} & 325 & - & 77.2 & -  & 25 \\
         NSENet \cite{nsenet} & 333 & 7.6 & 77.3 & -  & 167  \\
         GreedyNASv2-S \cite{greedynasv2} & 324 & 5.7 & 77.5 & 93.5 & 7 \\
         Cream-S \cite{cream} & 287 & 6.0 & 77.6 & 93.3 & 12 \\
         \textbf{GENIUS - 329 (Ours)} & 329 & 7.0 & \textbf{77.8} & \textbf{93.7}  & 5.6  \\ \midrule
         ProxylessNAS \cite{proxylessnas} & 465 & 7.1 & 75.1 & - & 15 \\
         SCARLET-A \cite{scarletnas} & 365 & 6.7 & 76.9 & 93.4 & 24 \\
         GreedyNAS-A  \cite{greedynas} & 366 & 6.5 & 77.1 & 93.3 & 7 \\
         BossNet-M2 \cite{bossnas} & 403 & - & 77.4 & 93.6 & 10 \\
         DNA-B \cite{dna} & 403 & 4.9 & 77.5 & 93.3 & 8.5 \\
         EfficientNet-B0 (Timm) \cite{efficientnet, timm} & 390 & 5.3 & 77.7 & 93.3 & Unkown  \\
         ST-NAS-B \cite{st-nas} & 503 & 7.8 & 77.9 & 93.8 & Unkown \\
         MCT-NAS-A \cite{mcts} & 442 & 8.4 & 78.0 & \textbf{93.9} & 12 \\
         \textbf{GENIUS - 401 (Ours)} & 401 & 7.5 & \textbf{78.2} & 93.8 &  5.7 \\ 
         
        \bottomrule
    \end{tabular}
\end{table*}

\textbf{FLOPs Constraint}. In order to further limit the FLOPs of the generated architectures, we incorporate a FLOPs look-up table within the problem encoding, allowing the GPT-4 to efficiently access the FLOPs count for each operation in every layer. Despite this, we discovered that the resulting architectures frequently surpass the designated FLOPs constraints. To overcome this challenge, we introduce a supplementary iterative loop within the GENIUS framework, which involves evaluating the actual FLOPs, providing feedback to the GPT-4, and iteratively refining the design until it adheres to the FLOPs constraint.

\textbf{Retraining Strategy}. Upon selecting the optimal architecture from the validation set, we adhere to the training strategy employed in a majority of prior works. \cite{efficientnet, greedynas, cream}. Specifically, we utilize the standard SGD optimizer with a momentum of 0.9 and a weight decay of 4e-5. We set an initial learning rate of 0.8, incorporating a warmup period of 5 epochs, and apply the cosine learning rate scheduler as described in \cite{cosine_lr}. Additionally, we also involve a dropout rate of 0.2, RandAugment \cite{randaugment} with n=2 and m=9. Furthermore, we employ an exponential moving average (EMA) network, and the performance is reported on the EMA network. The retraining process is conducted using 8 NVIDIA A100 GPUs, with a batch size of 2,048 and 500 epochs.

\textbf{Results}. We conducted experiments using two popular FLOPs constraints, one at approximately 300M and another at around 400M. The results are presented in Table \ref{exp:imagenet}. Notably, our GENIUS architecture demonstrates strong performance when compared to previous work. 
In the 300M setting, GENIUS achieves a 77.8\% Top-1 accuracy with 329M FLOPs, which is 0.2\% higher than CREAM-S\cite{cream} at 287M FLOPs.
In the 400M setting, GENIUS attains a 78.2\% Top-1 accuracy with 401M FLOPs, outperforming MCT-NAS-A's 442M FLOPs by 0.2\%. 
These results validate the effectiveness of GENIUS.
We note that the search cost of GENIUS is significantly lower than that of previous methods. 
We also note, however, that under scenarios that multiple require architectures at different FLOP constraints, Single Path One Shot approaches allow the user to amortise the cost of the search over different FLOP constraints---this amortisation is not available to GENIUS in the simple formulation we propose.

\vspace{-2mm}
\subsection{Ablation Study}

\renewcommand\arraystretch{0.8}
\begin{wraptable}{r}{6.2cm}
    \centering
    \vspace{-7mm}
    \caption{\footnotesize Training epochs for search stage. We fix input size as 224 in this experiment.}
    \label{exp:epoch}
    \footnotesize
    \setlength\tabcolsep{4pt}
    \begin{tabular}{c c c c c c}
        \toprule
        Epochs & 10 & 20 & 30 & 40 & 50 \\ \midrule
        Top-1 (\%) & 76.4 & \textbf{76.7} & 76.7 & 76.7 & 76.6 \\
        Search Cost & 3.0 & \textbf{6.1} & 9.2 & 12.3 & 15.4 \\
        \bottomrule
    \end{tabular}

    \caption{\footnotesize Input size for search stage. We fix epochs as 20 in this experiment.}
    \label{exp:input_size}
    \footnotesize
    \setlength\tabcolsep{4pt}
    \begin{tabular}{c c c c c c}
        \toprule
        Input Size & 224 & 196 & 160 & 128 & 96 \\ \midrule
        Top-1 (\%) & 76.7 & \textbf{76.9} & 76.7 & 76.5 & 76.2 \\
        Search Cost & 6.1 & \textbf{5.6} & 5.2 & 4.8 & 4.5 \\
        \bottomrule
    \end{tabular}
    \vspace{-2mm}
\end{wraptable}
In NAS, search cost is a primary concern. Our GENIUS method necessitates providing the actual accuracy of a model as feedback to GPT-4. For large-scale datasets (e.g., ImageNet), executing a standard training strategy for each generated architecture can be exceedingly costly. Instead, we can train the model in a lower-cost setting, as long as the results are adequately informative for GPT-4. Consequently, we conduct two experiments, as shown in Table \ref{exp:epoch} and \ref{exp:input_size}, to investigate the trade-off between search cost and final accuracy. We constrain the FLOPs to approximately 300M in this experiment and select the best architecture based on 1\% validation accuracy. The retraining strategy remains the same as in previous experiments, but with only 180 epochs.

Table \ref{exp:epoch} demonstrates the results of training the GENIUS-suggested model for 10, 20, 30, 40, and 50 epochs, respectively, and providing feedback to GPT-4. We observe that 20 epochs of training yields sufficient accuracy to supply informative feedback; additional epochs do not improve the final accuracy but substantially increase the search cost. In Table \ref{exp:input_size}, we fix the training epochs at 20 and vary the input size: 224, 196, 160, 128, and 96. Optimal results are achieved with an input size of 196, while further reducing the input size considerably diminishes the final accuracy. Therefore, our default setting consists of 20 epochs of training with an input size of 196. We recognize that GPT-4's response variability might influence the ablation study. However, given the stability of the results in these two experiments, we believe they still provide valuable insights into the trade-off between search cost and final accuracy.

\subsection{Transfer Learning}
\textbf{Classification}. We further assess the transferability of our GENIUS model by employing an ImageNet-pretrained model, which we subsequently fine-tune on the CIFAR10 and CIFAR100 datasets. The experimental setup adheres to the procedures outlined in \cite{fairnas, Gpipe}. Table \ref{exp:transfer_classification} presents the results, indicating that GENIUS achieves marginally superior performance compared to FairNAS-A \cite{fairnas}, thereby demonstrating its transferability.

\renewcommand\arraystretch{0.8}
\begin{table*}[h]
    \centering
    \vspace{-5mm}
    \caption{\footnotesize Transfer learning performance on the classification task.}
    \label{exp:transfer_classification}
    \footnotesize
    \setlength\tabcolsep{4pt}
    \begin{tabular}{c c c c c c}
        \toprule
        Backbone & Input Size & FLOPs(M) & Param(M) & CIFAR-10 & CIFAR-100 \\ \midrule
        NASNet-A \cite{nas2} & 331 $\times$ 331 & 12030 & 85 & 98.0 & 86.7 \\
        EfficientNet-B0 \cite{efficientnet} & 224 $\times$ 224 & 387 & 5.3 & 98.1 & 86.8 \\
        MixNet-M  \cite{mixconv} & 224 $\times$ 224 & 359 & 5.0 &  97.9 & 87.1 \\
        FairNas-A  \cite{fairnas} & 224 $\times$ 224 & 391 & 5.9 & 98.2 & 87.3 \\
        FairNas-C  \cite{fairnas} & 224 $\times$ 224 & 324 & 5.6 & 98.0 & 86.7 \\ \midrule
        \textbf{GENIUS - 329 (Ours)}  & 224 $\times$ 224 & 329 & 7.0 & 98.2 & 87.3 \\
        \textbf{GENIUS - 401 (Ours)}  & 224 $\times$ 224 & 401 & 7.5 & \textbf{98.3} & \textbf{87.4} \\
        \bottomrule
    \end{tabular}
    \vspace{-2mm}
\end{table*}

\textbf{Object Detection}. In this section, we assess the performance of our model by applying it to the object detection task, following the experimental setup detailed in \cite{fairnas}. Specifically, we directly employ the configuration file\footnote{\url{https://github.com/open-mmlab/mmdetection/blob/main/configs/retinanet/retinanet_r50_fpn_1x_coco.py}} from MMDetection \cite{mmdetection} and override the model definitions. We train the model on the MS COCO \cite{coco} \textit{train2017} set (118k images) and evaluate it on the \textit{val2017} set (5k images). The model is optimized over 12 epochs using a batch size of 16 across 8 GPUs. The initial learning rate is set to 0.01 and is decayed by a factor of 0.1 at epochs 8 and 11. We use the SGD optimizer with a momentum of 0.9 and weight decay of 1e-4. The detection algorithm employed in this study is RetinaNet \cite{retinanet}. Performance results are presented in Table \ref{exp:detection}, demonstrating that our GENIUS model significantly outperforms previous methods in the detection task. For instance, our model achieves a 7.8\% improvement over the MobileNetV2 baseline and a 0.9\% enhancement compared to prior art (GreedyNASV2). These experiments indicate that the GENIUS-searched architectures possess strong generalization capabilities in localization-sensitive tasks.

\renewcommand\arraystretch{0.8}
\begin{table*}[h]
    \centering
    \vspace{-2mm}
    \caption{\footnotesize Transfer learning performance on object detection with RetinaNet \cite{retinanet}. The performance is evaluated on COCO \textit{val2017}. The FLOPs are evaluated with 224x224 inputs. Top-1 indicates the ImageNet Results. The results for other methods are directly copied from \cite{cream}.}
    \label{exp:detection}
    \footnotesize
    \begin{tabular}{c c c c c c c c c}
        \toprule
         Backbone & FLOPs(M) & AP(\%) & AP$_{50}$ & AP$_{75}$ & AP$_{S}$ & AP$_{M}$ & AP$_{L}$ & Top-1 \\ \midrule
         MobileNetV2 \cite{mbv2} & 300 & 28.3 & 46.7 & 29.3 & 14.8  & 30.7 & 38.1 & 72.0 \\
         MobileNetV3 \cite{mbv3} & 219 & 29.9 & 49.3 & 30.8 & 14.9 & 33.3 & 41.1 & 75.2 \\
         MnasNet-A2 \cite{mnasnet} & 340 & 30.5 & 50.2 & 32.0 & 16.6 & 34.1 & 41.1 & 75.6 \\
         SPOS \cite{singlepath} & 365 & 30.7 & 49.8 & 32.2 & 15.4 & 33.9 & 41.6 & 75.0 \\
         FairNAS-C \cite{fairnas} & 325 & 31.2 & 50.8 & 32.7 & 16.3 & 34.4 & 42.3 & 76.7 \\
         MixNet-M \cite{mixconv} & 360  & 31.3 & 51.7 & 32.4 & 17.0 & 35.0 & 41.9 & 77.0 \\
         MixPath-A \cite{mixpath} & 349 & 31.5 & 51.3 & 33.2 & 17.4 & 35.3 & 41.8 & 76.9 \\
         FairNAS-A \cite{fairnas} & 392 & 32.4 & 52.4 & 33.9 & 17.2 & 36.3 & 43.2 & 77.5 \\
         Cream-S \cite{cream} & 287 & 33.2 & 53.6 & 34.9 & 18.2 & 36.6 & 44.4 &  77.6 \\ 
         GreedyNASV2 \cite{greedynasv2} & 324 & 34.9 & - & - & - & - & - & 77.5 \\ \midrule
         \textbf{GENIUS - 329 (Ours)} & 329 & \textbf{35.8} & \textbf{55.3} &  \textbf{38.1} & \textbf{19.6} & \textbf{38.7}  & \textbf{48.1} & \textbf{77.8} \\
        \bottomrule
    \end{tabular}
    \vspace{-2mm}
\end{table*}

\subsection{Eliciting design principles from GPT-4}
The architectures uncovered by GENIUS exhibit strong performance on various visual tasks, surpassing the state-of-the-art in several cases. 
Moreover, our search required little domain expertise---instead, this technical burden was transferred to GPT-4.
It is therefore of interest to examine how GPT-4 describes the principles by which it tackles the search problem.
When prompted, it offers the following maxims when considering the MobileNetV2 search space:
\textit{(1) In the initial stages, employ simpler operations to effectively capture low-level information. Subsequently, integrate more complex operations in the later stages to accurately represent higher-level information.
(2) To capture intricate features, the latter stages of the network must exhibit increased depth, while the earlier stages can maintain a shallower architecture.}
We have no guarantee that these introspective descriptions truly reflect the principles used during the GENIUS search process~\cite{turpin2023language}, but they nevertheless suggest an interpretable, intuitive characterisation of how GPT-4 approaches a specific search scenario.

%% file: section/limitation.tex
\section{Limitations}
\label{sec:limitations}

We identify several important limitations to our study.

\textbf{Reproducibility.} First, we have little insight into the operations that wrap GPT-4 inference behind the API provided by OpenAI.
For example, we do not know if our problem encoding text is pre-processed or if the model response is post-processed in some way (for example, by content moderation policies that are opaque to API clients).
It is possible that any such operations change over the course of an experiment, and we are unable to control for such changes.
Second, even with the temperature set to 0, we observe some variation in GPT-4 responses, making it challenging to numerically reproduce a particular experimental run. 

\textbf{Benchmark contamination.}
We do not know which data was included in the training set for GPT-4, or the final cut-off date for training data provided to the model\footnote{In~\cite{gpt4}, the authors note: \textit{GPT-4 generally lacks knowledge of events that have occurred after the vast majority of its pre-training data cuts off in September 2021.... the pre-training and post-training data contain a small amount of more recent data.}}.
It is therefore possible that the benchmarks employed in our studies have all been ``seen'' by GPT-4, and thus it is searching ``from memory'' rather than leveraging insight about how to improve an architecture design. 
We note that previous studies examining the evidence of contamination have often found its effect on final performance to be somewhat limited~\cite{brown2020language,radford2021learning}, perhaps due to the challenge of memorizing so much magnitude of the training data. 
Nevertheless, the fact that we cannot rule out contamination represents a significant caveat to our findings.
One potential solution to address this in future work could be the construction of private optimisation benchmarks that are hidden from the open internet to ensure that they are excluded from the training data of large language models.

\textbf{Limited control and inscrutability.}
Prompting represents our sole point of control over GPT-4, but we have relatively little understanding of how changes to the prompt influence behaviour as an optimiser.
On the NAS-Bench-201 benchmark (see more details in Appendix A.3.), we find that later iterations under-perform earlier iterations in some cases, and it is unclear why this should be the case given that: (i) our prompt requests improved performance, (ii) our experimental evidence suggests that GPT-4 is capable of providing improved performance. 
We believe future work on this problem is particularly valuable.

%% file: section/safety.tex
\section{AI safety}
\label{sec:ai_safety}

As AI systems become more capable, they exhibit greater potential for useful applications.
However, they also represent greater risk---a concern that has been discussed by leading researchers within the field of AI for more than 60 years~\cite{Wiener1960SomeMA}.
The use of GPT-4 as a black-box optimiser can potentially represent an offloading of intellectual labour from a human researcher to an inscrutable system.
This contributes to the risk of \textit{enfeeblement}~\cite{hendrycks2022x} in which know-how erodes by delegating increasingly many important functions to machines. 
If general-purpose black-box optimisers ultimately prove superior to interpretable alternatives, competition pressures may incentivise such delegation~\cite{hendrycks2023natural}.
Architecture search, in particular, represents a potential vector for self-improvement (potentially complementing strategies that improve the inference capabilities of a trained model~\cite{huang2022large}).
Such research can yield improved performance on tasks deemed beneficial by society, but may also exacerbate risk.

We believe it is useful to study whether existing, publicly available frontier models like GPT-4 possess such capabilities.
Our tentative results (subject to the important limiting caveats described in Sec.~\ref{sec:limitations}), taken together with concurrent studies of scientific automation in other domains~\cite{bran2023chemcrow,boiko2023emergent}, suggest that GPT-4 could potentially represent an artefact that leads to accelerated scientific research and therefore caution is appropriate in its application. %

%% file: section/conclusion.tex
\section{Conclusion}
In this paper, we present GENIUS, a novel NAS approach that employs the GPT-4 language model as a black-box optimiser to expedite the process of discovering efficient neural architectures. 
We compare GENIUS against leading NAS methods, underscoring its effectiveness and highlighting the of GPT-4 as a tool for research and development. 
We also note safety implications and discuss several important limitations of our work.
In future work, we plan to further study the capabilities and limitations of GPT-4 (and other frontier language models) to serve as optimisers in applications that have traditionally required extensive domain expertise, and to more extensively investigate the safety implications of such research.

%% file: section/acknowledgements.tex
\section*{Acknowledgment}
We thank H. Bradley for suggesting the method acronym.
This work was supported by an Isaac Newton Trust grant.
S. Albanie would like to thank Z. Novak and N. Novak for enabling his contribution.

%% file: section/appendixv2.tex
\section{Numerical Results}

\subsection{Detailed Numerical Results for Figure \ref{fig:nas_bench_macro}}
\renewcommand\arraystretch{0.8}
\begin{table*}[h]
    \centering
    \caption{\footnotesize Experimental Results on NAS-Bench-Macro. We set \textbf{Temperature = 1} for GPT-4 in this experiment. T is the iteration.}
    \label{table:nas_bench_macro_t1}
    \footnotesize
    \setlength\tabcolsep{4pt}
    \begin{tabular}{c  c c c c c c c c c c c | c}
        \toprule
         & & T = 0 &  T = 1 & T = 2 & T = 3 & T = 4 & T = 5 & T = 6 & T = 7 & T = 8 & T = 9 & Optimal \\ \midrule
         \multirow{2}{*}{Trial 1} & Acc  & 90.90 & 92.40 & 92.30 & 92.53 & 92.63 & 92.66 & \textbf{92.97} & 92.56 & 92.50 & 92.56 & \textcolor{gray}{93.13} \\
         & Ranking & 3440 & 590 & 766 & 353 & 203 & 180 & \textbf{19} & 311 & 394 & 314 & \textcolor{gray}{1} \\ \midrule
         \multirow{2}{*}{Trial 2} & Acc & 90.42 & 92.49 & 92.53 & \textbf{92.85} & 92.54 & 92.56 & 92.58 & 92.73 & 92.48 & 92.78 & \textcolor{gray}{93.13} \\
         & Ranking & 4042 & 442 & 384 & \textbf{50} & 332 & 331 & 272 & 119 & 446 & 82 & \textcolor{gray}{1}\\ \midrule
         \multirow{2}{*}{Trial 3} & Acc  & 91.35 & 92.78 & \textbf{92.82} & 92.74 & 92.34 & 92.35 & 92.45 & 92.56 & 92.54 & 92.66 & \textcolor{gray}{93.13} \\
         & Ranking & 2609 & 83 & \textbf{65}  & 117 & 683 & 664 & 483 & 311 & 341 & 180 & \textcolor{gray}{1} \\
        \bottomrule
    \end{tabular}
    \vspace{-3mm}
\end{table*}
\renewcommand\arraystretch{0.8}
\begin{table*}[h]
    \centering
    \caption{\footnotesize Experimental Results on NAS-Bench-Macro. We set \textbf{Temperature = 0} for GPT-4 in this experiment. T is the iteration.. `-' denotes that GPT-4 asserts there is no chance to improve the performance further.}
    \label{table:nas_bench_macro_t0}
    \footnotesize
    \setlength\tabcolsep{4pt}
    \begin{tabular}{c  c c c c c c c c c c c | c}
        \toprule
         & & T = 0 &  T = 1 & T = 2 & T = 3 & T = 4 & T = 5 & T = 6 & T = 7 & T = 8 & T = 9 & Optimal \\ \midrule
         \multirow{2}{*}{Trial 1} & Acc  & 85.70 & 92.62 & 92.82 & \textbf{93.05} & 92.95 & 92.46 & - & - & - & - & \textcolor{gray}{93.13} \\
         & Ranking & 6221 & 212 & 64 & \textbf{8} & 21 & 479 & - & - & - & - & \textcolor{gray}{1} \\ \midrule
         \multirow{2}{*}{Trial 2} & Acc  & 92.45  & 92.66 & \textbf{92.92} & 92.64 & 92.33 & 92.72 & - & - & - & - & \textcolor{gray}{93.13} \\
         & Ranking & 496 & 189 & \textbf{27} & 198 & 695 & 128 & - & - & - & - & \textcolor{gray}{1} \\ \midrule
         \multirow{2}{*}{Trial 3} & Acc  & 92.41 & 92.74 & \textbf{92.83} & 92.74 & 92.33 & 92.53 & 92.69 & 92.34 & 92.56 & 92.72 & \textcolor{gray}{93.13} \\
         & Ranking & 564 & 113 & \textbf{61} & 112 & 689 & 352 & 152 & 683 & 314 & 128 & \textcolor{gray}{1}\\
        \bottomrule
    \end{tabular}
\end{table*}

\subsection{Detailed Numerical Results for Figure \ref{fig:channel_bench_macro}}

\renewcommand\arraystretch{0.8}
\begin{table*}[!h]
    \centering
    \setlength\tabcolsep{4pt}
    \caption{\footnotesize Experimental Results on Channel-Bench-Macro with \textbf{ResNet}. We set \textbf{Temperature = 0} for GPT-4 in this experiment. T is the iteration.}
    \label{table:channel_bench_macro_res}
    \footnotesize
    \begin{tabular}{c c c c c c c c c c c c | c}
        \toprule
         & & T = 0 &  T = 1 & T = 2 & T = 3 & T = 4 & T = 5 & T = 6 & T = 7 & T = 8 & T = 9 & Optimal \\ \midrule
         \multirow{2}{*}{Trial 1} & Acc  & 93.06 & 93.52 & 93.56 & 93.61 & 93.70 & 93.73 & 93.62 & \textbf{93.76} & 93.49 & 93.66 & \textcolor{gray}{93.89}\\
         & Ranking & 9862 & 1205 & 737  & 411 &  103 & 61 & 365 & \textbf{33} & 1515 & 173 & \textcolor{gray}{1}\\ \midrule
         \multirow{2}{*}{Trial 2} & Acc  & 92.89 & 93.41 & 93.65 & 93.64 & 93.67 & 93.48 & \textbf{93.82} & 93.62 & 93.80 & 93.64 & \textcolor{gray}{93.89}\\
         & Ranking & 12457 & 2813 & 209 & 272 & 142 & 1708 & \textbf{8} & 376 & 13 & 260 & \textcolor{gray}{1}\\ \midrule
         \multirow{2}{*}{Trial 3} & Acc  & 92.60 & 93.41 & 93.44 & 93.64 & 93.65 & 93.67 & \textbf{93.75} & 93.62 & 93.66 & 93.73 & \textcolor{gray}{93.89}\\
         & Ranking & 15178 & 2813 & 2349 & 272 & 190 & 142 & \textbf{37} & 376 & 181 & 59 & \textcolor{gray}{1}\\ 
        \bottomrule
    \end{tabular}
\end{table*}
\renewcommand\arraystretch{0.8}
\begin{table*}[!h]
    \centering
    \setlength\tabcolsep{4pt}
    \caption{\footnotesize Experimental Results on Channel-Bench-Macro with \textbf{MobileNet}. We set \textbf{Temperature = 0} for GPT-4 in this experiment. T is the iteration. `-' denotes that GPT-4 asserts there is no chance to improve the performance further.}
    \label{table:channel_bench_macro_mob}
    \footnotesize
    \begin{tabular}{c c c c c c c c c c c c | c}
        \toprule
         & & T = 0 &  T = 1 & T = 2 & T = 3 & T = 4 & T = 5 & T = 6 & T = 7 & T = 8 & T = 9 & Optimal \\ \midrule
         \multirow{2}{*}{Trial 1} & Acc  & 91.02 & 91.46 & 91.66 & \textbf{91.76} & 91.35 & 91.63 & 91.53 & 91.35 & 91.46 & 91.51 & \textcolor{gray}{91.89}\\
         & Ranking & 5383  & 630   &  65 & \textbf{16} & 1383 & 80 & 318 & 1383 & 630 & 380 & \textcolor{gray}{1} \\ \midrule
         \multirow{2}{*}{Trial 2} & Acc  & 91.03 & 91.60 & 91.66 & \textbf{91.76} & 91.42 & 91.53 & 91.67 & \textbf{91.76} & - & - & \textcolor{gray}{91.89}\\
         & Ranking & 5271 & 146 & 65 & \textbf{16} & 871 & 318 & 59 & \textbf{16} & - & - & \textcolor{gray}{1}\\ \midrule
         \multirow{2}{*}{Trial 3} & Acc & 91.07 & 91.52 & 91.66 & 91.59 & \textbf{91.76} & 91.66 & 91.42 & 91.67 & \textbf{91.76} & - & \textcolor{gray}{91.89} \\ 
         & Ranking  & 4707 & 359 & 65 & 164 & \textbf{16} & 65 & 871 & 59 & \textbf{16} & - & \textcolor{gray}{1} \\
        \bottomrule
    \end{tabular}
\end{table*}

\newpage

\subsection{Detailed Numerical Results for NAS-Bench-201}

\renewcommand\arraystretch{0.8}
\begin{table*}[h]
    \centering
    \caption{\footnotesize Experimental Results on NAS-Bench-201 with \textbf{CIFAR10}. We set Temperature = 0 for GPT-4 in this experiment. T is the iteration. We perform GENIUS on the validation set and report the final accuracy and ranking on the test set based on the best architectures verified on the validation set.}
    \label{table:nas_bench_201_cifar10}
    \footnotesize
    \setlength\tabcolsep{4pt}
    \begin{tabular}{c  c c c c c c c c c c | c c}
        \toprule
         & T = 0 &  T = 1 & T = 2 & T = 3 & T = 4 & T = 5 & T = 6 & T = 7 & T = 8 & T = 9 & Test Acc & Ranking \\ \midrule
         Trial 1  & 90.59 & 90.15 & 90.42 & 90.06 & 89.60 & \textbf{91.28} & 90.15 & 90.92 & 90.88 & 86.41 & 93.79 & 142 \\ 
         Trial 2  & \textbf{90.88} & 90.44 & 90.86 & 88.99 & 87.45 & 90.05 & 89.33 & 88.65 & 90.43 & 85.10 & 93.83 & 114 \\ 
         Trial 3  & \textbf{91.28} & 90.80 & 90.88 & 91.10 & 90.15 & 86.41 & 86.45 & 86.34 & 86.17 & 86.33 & 93.79 & 142 \\
         Trial 4 & 90.60 & 90.05 & 90.43 & 89.36 & 89.93 & 90.05 & \textbf{91.01} & 90.85 & 89.36 & 88.64 & 93.92 & 62 \\
         Trial 5 & 90.36 & 89.48 & 89.52 & 89.98 & \textbf{90.80} & 89.90 & 89.36 & 89.28 & 89.82 & 89.98 & 93.64 & 279 \\
         \bottomrule
    \end{tabular}
\end{table*}

\renewcommand\arraystretch{0.8}
\begin{table*}[h]
    \centering
    \setlength\tabcolsep{4pt}
    \caption{\footnotesize Experimental Results on NAS-Bench-201 with \textbf{CIFAR100}. We set Temperature = 0 for GPT-4 in this experiment. T is the iteration. We perform GENIUS on the validation set and report the final accuracy and ranking on the test set based on the best architectures verified on the validation set.}
    \label{table:nas_bench_201_cifar100}
    \footnotesize
    \begin{tabular}{c  c c c c c c c c c c | c c}
        \toprule
         & T = 0 &  T = 1 & T = 2 & T = 3 & T = 4 & T = 5 & T = 6 & T = 7 & T = 8 & T = 9 & Test Acc & Ranking \\ \midrule
         Trail 1 & 69.46 & \textbf{71.29} & 69.66 & 69.46 & 68.79 & 70.75 & 69.16 & 70.24 & 70.16 & 69.18 & 71.51 & 103 \\
         Trial 2  & 70.05 & 65.33 & 69.64 & \textbf{70.81} & 66.77 & 65.19 & 70.70 & 70.37 & 66.85 & 65.84 & 70.78 & 292 \\
         Trial 3  & 69.39 & 69.56 & 69.43 & \textbf{70.65} & 70.35 & 68.04 & 69.59 & 67.64 & 67.97 & 66.44 & 70.16 & 724 \\
         Trail 4 & 70.62 & 65.53 & \textbf{71.42} & 70.78 & 69.56 & 65.77 & 70.78 & 68.88 & 65.53 & 66.23 & 71.96 & 57 \\
         Trail 5 & 67.00 & 70.35 & 69.59 & 65.05 & 67.17 & 65.34 & \textbf{70.65} & 70.35 & 69.16 & 68.31 & 70.16 & 724 \\
         \bottomrule
    \end{tabular}
\end{table*}

\renewcommand\arraystretch{0.8}
\begin{table*}[h]
    \centering
    \setlength\tabcolsep{4pt}
    \caption{\footnotesize Experimental Results on NAS-Bench-201 with \textbf{ImageNet16-120}. We set Temperature = 0 for GPT-4 in this experiment. T is the iteration. We perform GENIUS on the validation set and report the final accuracy and ranking on the test set based on the best architectures verified on the validation set.}
    \label{table:nas_bench_201_imagenet}
    \footnotesize
    \begin{tabular}{c  c c c c c c c c c c | c c}
        \toprule
         & T = 0 &  T = 1 & T = 2 & T = 3 & T = 4 & T = 5 & T = 6 & T = 7 & T = 8 & T = 9 & Test Acc & Ranking \\ \midrule

         Trial 1 & 44.78 & 45.25 & 44.17 & \textbf{46.40} & 46.27 & 46.32 & 44.20 & 42.77 & 44.67 & 43.96 & 46.67 & 8 \\
         Trial 2 & 43.47 & 43.85 & \textbf{45.57} & 45.13 & 44.83 & 40.08 & 45.23 & 40.24 & 45.27 & 43.15 & 45.51 & 135 \\
         Trial 3 & 36.92 & 43.95 & 38.06 & \textbf{44.23} & 39.87 & 40.23 & 43.87 & 42.00 & 38.38 & 43.53 & 44.00 & 813 \\
         Trial 4 & 42.64 & 40.77 & 38.73 & \textbf{44.48} & 44.29 & 39.96 & 43.23 & 44.11 & 39.96 & 44.39 & 43.96 & 850 \\
         Trial 5 & \textbf{45.75} & 43.95 & 42.00 & 40.23 & 43.27 & 43.03 & 39.667 & 43.95 & 42.00 & 43.60 & 44.65 & 434 \\
         \bottomrule
    \end{tabular}
\end{table*}

\section{ImageNet}

\subsection{MobileNetV2 Search Space}
\renewcommand\arraystretch{0.8}
\begin{table}[!h]
\centering
\footnotesize
\caption{The "Num Blocks" represents the maximum number of blocks in a group. The "Stride" indicates the convolutional stride of the first block in each group.}
\label{tab:search_space}
\begin{tabular}{ccccc}
    \toprule
    Input Shape & Operators & Channels & Num Blocks & Stride \\ \midrule
    $224^2\times3$  & $3\times3$ Conv    & 16    & 1 & 2\\
    $112^2\times16$  & $3\times3$ Depthwise \& Pointwise Conv    & 16    & 1 & 2\\ \midrule
    $56^2\times16$ & \textbf{Choice Block}         & 24   & 6 & 2\\
    $28^2\times24$ & \textbf{Choice Block}         & 40   & 6 & 2\\
    $14^2\times40$ & \textbf{Choice Block}         & 80   & 6 & 1\\
    $14^2\times80$ & \textbf{Choice Block}         & 96   & 6 & 2\\
    $7^2\times96$  & \textbf{Choice Block}          & 192   & 6 & 1\\ \midrule
    $7^2\times192$  &  $1\times1$ Conv   & 320  & 1 & 1\\
    $7^2\times320$  & Global Avg. Pooling    & 320  & 1 & 1\\
    $320$  & $1\times1$ Conv    & 1,280  & 1 & 1\\
    $1,280$ & Fully Connect & 1,000 & 1 & -\\
    \bottomrule
\end{tabular}
\end{table}

\newpage
\subsection{Architecture Details}

\textbf{GENIUS - 329}
\begin{lstlisting}
# Input 56 x 56 x 16
InvertedResidual(kernel_size=3, exp_ratio=4)
InvertedResidual(kernel_size=3, exp_ratio=4)

# Input 28 x 28 x 24
InvertedResidual(kernel_size=3, exp_ratio=4)
InvertedResidual(kernel_size=3, exp_ratio=4)
InvertedResidual(kernel_size=3, exp_ratio=4)
InvertedResidual(kernel_size=3, exp_ratio=4)

# Input 14 x 14 x 40
InvertedResidual(kernel_size=3, exp_ratio=6)
InvertedResidual(kernel_size=3, exp_ratio=6)
InvertedResidual(kernel_size=7, exp_ratio=4)

# Input 14 x 14 x 80
InvertedResidual(kernel_size=3, exp_ratio=6)
InvertedResidual(kernel_size=7, exp_ratio=4)
InvertedResidual(kernel_size=3, exp_ratio=6)
InvertedResidual(kernel_size=3, exp_ratio=6)

# Input 7 x 7 x 96
InvertedResidual(kernel_size=5, exp_ratio=6)
InvertedResidual(kernel_size=5, exp_ratio=6)
InvertedResidual(kernel_size=5, exp_ratio=6)
InvertedResidual(kernel_size=5, exp_ratio=6)
\end{lstlisting}

\textbf{GENIUS - 401}
\begin{lstlisting}
# Input 56 x 56 x 16
InvertedResidual(kernel_size=3, exp_ratio=4)
InvertedResidual(kernel_size=3, exp_ratio=4)
InvertedResidual(kernel_size=3, exp_ratio=4)
InvertedResidual(kernel_size=3, exp_ratio=4)

# Input 28 x 28 x 24
InvertedResidual(kernel_size=3, exp_ratio=4)
InvertedResidual(kernel_size=3, exp_ratio=4)
InvertedResidual(kernel_size=3, exp_ratio=4)
InvertedResidual(kernel_size=3, exp_ratio=4)

# Input 14 x 14 x 40
InvertedResidual(kernel_size=3, exp_ratio=6)
InvertedResidual(kernel_size=3, exp_ratio=6)
InvertedResidual(kernel_size=3, exp_ratio=6)
InvertedResidual(kernel_size=3, exp_ratio=6)

# Input 14 x 14 x 80
InvertedResidual(kernel_size=3, exp_ratio=6)
InvertedResidual(kernel_size=5, exp_ratio=6)
InvertedResidual(kernel_size=5, exp_ratio=6)
InvertedResidual(kernel_size=5, exp_ratio=6)

# Input 7 x 7 x 96
InvertedResidual(kernel_size=7, exp_ratio=6)
InvertedResidual(kernel_size=5, exp_ratio=6)
InvertedResidual(kernel_size=7, exp_ratio=6)
InvertedResidual(kernel_size=5, exp_ratio=6)
\end{lstlisting}